\documentclass[acmsmall,authorversion,nonacm]{acmart}
\AtBeginDocument{%
  \providecommand\BibTeX{{%
    \normalfont B\kern-0.5em{\scshape i\kern-0.25em b}\kern-0.8em\TeX}}}

\usepackage{booktabs}
\usepackage{algorithm}
\usepackage{algpseudocode}
\usepackage{tikz}
\usepackage{amsmath}
\usepackage{caption}
\usepackage{cleveref}
\usepackage{wrapfig}
\usepackage{booktabs}
\usepackage{newfloat}
\usepackage{listings}
\DeclareCaptionStyle{ruled}{labelfont=normalfont,labelsep=colon,strut=off} % DO NOT CHANGE THIS
\floatstyle{ruled}
\newfloat{listing}{tb}{lst}{}
\floatname{listing}{Listing}

\setcopyright{none}
\begin{document}


% \title{Predicting Depression and Anxiety at the Onset of COVID-19: An Artificial Neural Network Approach}
\title{Predicting Depression and Anxiety: A Multi-Layer Perceptron for Analyzing the Mental Health Impact of COVID-19}

\author{David Fong}
\email{fongd2@rpi.edu}

\author{Tianshu Chu}
\email{chut@rpi.edu}

\author{Matthew Heflin}
\email{matthew.heflin@mssm.edu}

\author{Xiaosi Gu}
\email{xiaosi.gu@mssm.edu}

\author{Oshani Seneviratne}
\email{senevo@rpi.edu}

\begin{abstract}
% OS: Rewrote the abstract to make it concise:
We introduce a multi-layer perceptron (MLP) called the COVID-19 Depression and Anxiety Predictor (CoDAP) to predict mental health trends, particularly anxiety and depression, during the COVID-19 pandemic. Our method utilizes a comprehensive dataset, which tracked mental health symptoms weekly over ten weeks during the initial COVID-19 wave (April to June 2020) in a diverse cohort of U.S. adults. This period, characterized by a surge in mental health symptoms and conditions, offers a critical context for our analysis. Our focus was to extract and analyze patterns of anxiety and depression through a unique lens of qualitative individual attributes using CoDAP. This model not only predicts patterns of anxiety and depression during the pandemic but also unveils key insights into the interplay of demographic factors, behavioral changes, and social determinants of mental health. These findings contribute to a more nuanced understanding of the complexity of mental health issues in times of global health crises, potentially guiding future early interventions.

\end{abstract}




\maketitle

\section{Introduction}
\label{sec:introduction}

Psychiatric disorders have always imposed significant challenges to public health ~\cite{prince2007no}, and the advent of the COVID-19 pandemic has intensified these issues~\cite{talevi2020mental,moreno2020mental,cullen2020mental}. The pandemic has led to increased prevalence and severity of mental health concerns globally, driven by a mix of health fears, social isolation, economic pressures, and general uncertainty. This environment presents new challenges and opportunities for understanding and addressing mental health conditions, in particular depression and anxiety.
% 
% OS: commented this paragraph, as we are not really talking about device data here.
% Proactive monitoring of mental health indicators, such as anxiety and depression levels, is more critical than ever. This monitoring provides mental health professionals with a richer, more objective understanding of their patients' conditions compared to traditional self-reporting methods. Such vigilance is crucial in tailoring treatments and interventions more effectively, ultimately improving patient outcomes. However, the practitioners should not be in a data deluge but instead be enabled to utilize automated tools for their diagnosis and treatment. 
% 
We employ an MLP to predict depression and anxiety, which can lead to early intervention and better management of such potential mental health conditions. Our approach has the unique ability to process complex data patterns in mental health datasets, making it particularly suitable for the multifaceted nature of mental health. Central to our study is a dataset comprising detailed weekly surveys~\cite{shuster2021emotional} that offers an invaluable perspective on the psychological impact of the pandemic. This data was collected at the beginning of the COVID-19 pandemic (April - June 2020) in the U.S. Even though the results of this study may not generalize well to pre- and post-pandemic times or other geographical locations, by analyzing this dataset, we gathered valuable insights into the mental health trends of periods during stressful global catastrophes such as the COVID-19 pandemic.

\subsection{Contribution}

We created a machine learning model designed to predict anxiety and depression levels using qualitative data from an individual during global health crises, such as the COVID-19 pandemic.  
% We can't say that because there's several papers on this same topic
% Our research stands out by applying MLP's to mental health prediction during a global crisis, a relatively unexplored area. 
We utilized model explainability to reveal specific qualitative attributes that increase susceptibility to such mental health issues, which is essential in guiding the creation of targeted mental health interventions, specifically addressing the needs of the most affected individuals. Our work aims to advance predictive modeling in mental health and deepen the understanding of psychological impacts during global health crises, offering valuable insights for practitioners and researchers.

% and an overview of mental resilience and emotional recovery after a period of hardship. % OS: I don't think we have any insights on recovery. So, commented that out. 

\section{Related Work}
\label{sec:related-work}

% ``Emotional adaptation during a crisis: Decline in anxiety and depression after the initial weeks of COVID-19 in the United States" (Shuster, 2021), was about a 10-week survey with a diverse group of participants, examining various demographic and psychological features, including anxiety and depression levels.  

The exploration of mental health, specifically anxiety and depression, in the context of the COVID-19 pandemic has been a focus of several studies. Our research, utilizing an MLP to decode patterns of depression and anxiety, aligns with these efforts but presents distinct methodologies and insights, as noted below.

% \subsection{Comparison with Machine Learning-Based Predictive Models}
\citeauthor{hueniken2021machine}~\cite{hueniken2021machine} utilized surveys to assess mental health symptoms, alongside factors like substance use and COVID-19-related worries, employing principal component analysis. Our work diverges from this work by focusing on extracting patterns specific to anxiety and depression using MLPs.
% 
% \subsection{Contrasting Coping Strategies and External Factors}
\citeauthor{freyhofer2021depression}~\cite{freyhofer2021depression} offer insights into the impact of coping strategies and external factors on mental health during the pandemic. In contrast, our research delves into qualitative attributes of individuals, using a neural network-based approach to uncover their interplay in mental health disorders.
% 
% \subsection{Methodological Insights from Nature's Publication}
\citeauthor{nemesure2021predictive}~\cite{nemesure2021predictive} provides methodological insights that could be compared to our MLP approach. While both aim at predictive modeling, our emphasis on quantitative data analysis through an MLP offers a novel perspective.
% 
% \subsection{Machine Learning in Detecting Depression}
\citeauthor{malik2023machine}~\cite{malik2023machine} discusses using various machine learning models for mental health analysis during COVID-19, using a combination of techniques such as decision trees, k-nearest neighbor, and naive Bayes. Our research contrasts this by predicting patterns of depression and anxiety, providing an alternate view of mental health dynamics.
% 
% \subsection{Longitudinal Data Analysis}
%  OS: commented since we did not do any longi
% \citeauthor{venanzi2022longitudinal} explores longitudinal predictors of mental health issues quantitatively. Our focus on interpreting patterns related to mental health conditions presents a distinct approach in analyzing longitudinal data, and we will be utilizing computational techniques for a more robust time series analysis.

In contrast to the data analysis methods described above, \citeauthor{shukla2023mentalhealthai}~\cite{shukla2023mentalhealthai} implemented a mood prediction system using personal medical devices that tracked patient physiological data. 
They build a decentralized learning machine learning model called ``MentalHealthAI" with federated learning and smart contracts to keep data on the patient's end without collecting it in a central server. This provides a privacy-preserving methodology for analyzing mental health data. 
% We drew inspiration from this paper through their work in mood prediction through various machine learning algorithms.
While ``MentalHealthAI"~\cite{shukla2023mentalhealthai} used a method for predicting mood through psychological data focusing on mental health as a chronic disease, our focus in this paper is on predicting depression and anxiety levels using qualitative survey data collected at the onset of a traumatic global health crisis such as the COVID-19 pandemic. 
% We also concentrated on analyzing the qualitative data of an individual to make the prediction rather than quantitative attributes. Through this, we aim to contribute a nuanced perspective to the predictive modeling of mental health.
% 
In a similar vein, \citeauthor{d2020ai}~\cite{d2020ai} focuses on three of the main ways that AI is applied in mental health: (1) personal sensing or digital phenotyping, (2) natural language processing of clinical texts and social media content, and (3) chatbots. These methodologies exemplify a promising future for integrating AI into digital interventions, especially within web and smartphone applications, and the authors also underscore the ethical challenges associated with using AI in mental health care.
% Our research was significantly influenced by the insights presented in the paper 'AI in Mental Health,' particularly regarding the ethical considerations surrounding AI and user privacy in mental health applications. Ultimately, 
Our goal is to develop an application that incorporates these ethical guidelines and advances the integration of AI in mental health services to assist mental health professionals.

\section{System Design and Implementation}
\label{sec:system-design}

\subsection{Dataset}
\label{sec:dataset}

Our work utilizes the dataset collected by \citeauthor{shuster2021emotional}~\cite{shuster2021emotional}, a 10-week survey with a diverse group of participants that examined 308 demographic and psychological features, including anxiety and depression levels. Participants represented a wide spectrum of diversity, coming from different ethnic backgrounds, genders, religious beliefs, socioeconomic statuses, geographical locations, and varying mental health conditions. \citeauthor{shuster2021emotional} leveraged linear mixed-effects models to examine factors contributing to longitudinal changes in depression and anxiety and discovered that gender, age, income level, previous psychiatric diagnosis, informedness about the pandemic, and marital status correlated with higher overall levels of anxiety and depression.

\begin{wrapfigure}{r}{0.5\textwidth} % 'r' for right side, and '0.5\textwidth' for the width of the wrap figure
  \centering
  \setlength{\abovecaptionskip}{2pt}
  \setlength{\belowcaptionskip}{2pt}
    % \vspace{-3ex}
  \includegraphics[width=\linewidth,trim={0 0 0 0},clip]{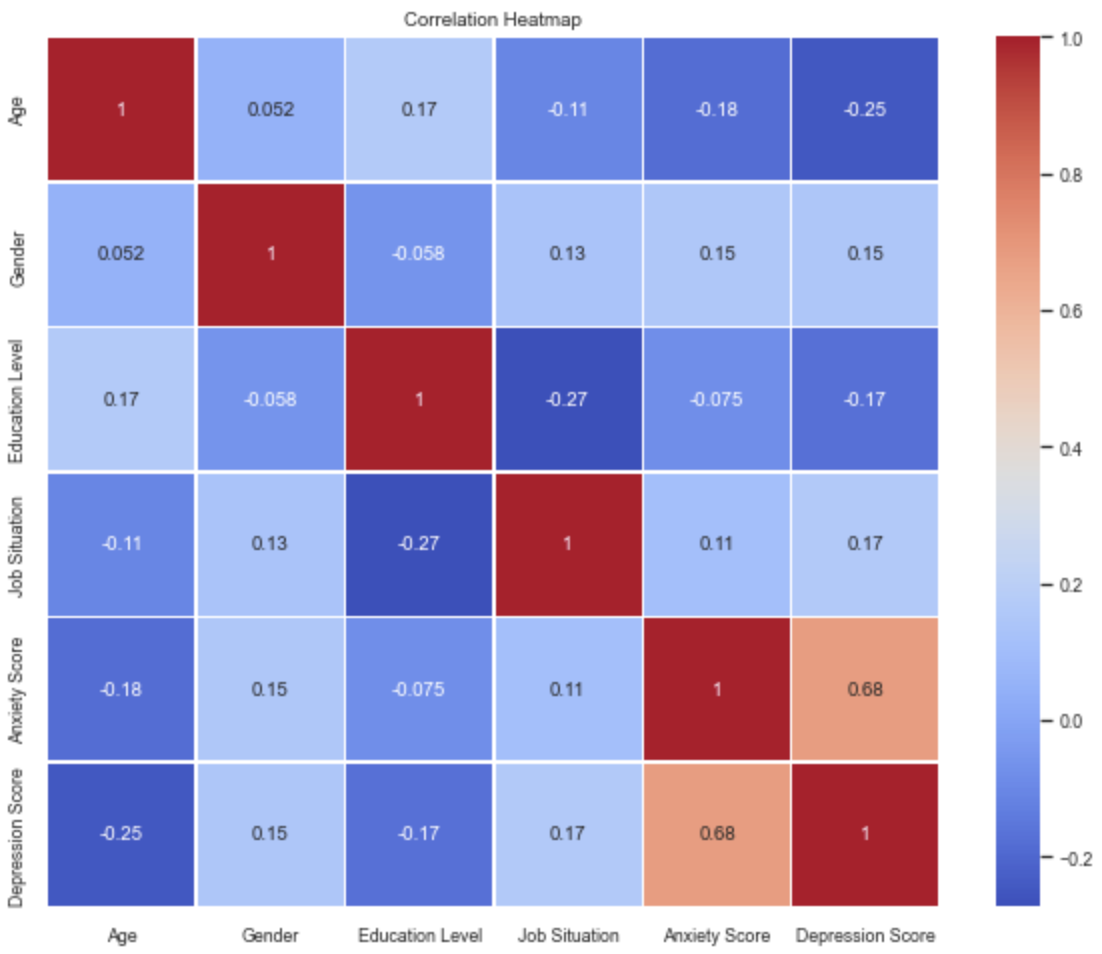} % trim to remove white space (left, bottom, right, top)
\caption{Heat map illustrating the relationships among four commonly examined personal attributes (age, gender, education level, job) and a primary metric of mental health conditions from the dataset by \citeauthor{shuster2021emotional}. The subtle variations in correlation strength underscore the challenge of manually discerning crucial features related to mental health conditions.}
  \label{fig:heatmap_correlation}
\vspace{-2ex}
\end{wrapfigure}

Using this same dataset, we created an MLP that predicts depression, i.e., \texttt{sds score}, as measured by the Self-Rating Depression Scale~\cite{zung1965self} and anxiety, i.e., \texttt{stai-s score}, as measured by the state sub-scale of the State-Trait Anxiety Inventory~\cite{spielberger1989state},  based on patient features. Our work extends and enhances the understanding gained from the foundational study that initially collected the data~\cite{shuster2021emotional}.
The dataset by \citeauthor{shuster2021emotional} consists of numeric and text data. The numeric data consisted of data from numeric or categorical answers, such as age, gender, race, employment status, mental health disorders, anxiety scores (\texttt{stai-s}), and depression scores (\texttt{sds}).
The textual data consists of user answers that were not categorical or numerical. One instance of these questions was where applicants had to describe various images they saw and what emotion they invoked. In our analysis, we only focused on the numerical data.

\subsection{Exploratory Data Analysis}
\label{sec:eda}

Data exploration is crucial for understanding the dataset's underlying structure and performing feature engineering. To ensure data integrity and consistency for our analyses, we followed a meticulous reformatting process involving several steps, such as selecting complete and consistent participant records, removing null data entries, and standardizing columns across the ten-week time points. 
An essential aspect of this phase was transforming categorical variables into a numerical format, enhancing their suitability for processing by our MLP.
After our exploratory data analysis, we split our dataset into a 70-20-10 for training, testing, and validation.

\subsection{Predictive Model}
\label{sec:model}
% Our predictive MLP model architecture is rooted in several compelling reasons. Firstly, the 
% The complexity of mental health data necessitates a model capable of discerning significant features from an extensive dataset. 

Mental health data, particularly concerning anxiety and depression, often involve intricate patterns and interactions among various factors such as demographic variables, behavioral patterns, and external stressors. 
Therefore, our choice of model revolved around deep learning-based MLPs, which are well-suited for modeling complex, nonlinear relationships within data, providing more accurate predictions compared to the traditional linear models used in \cite{shuster2021emotional}.
The MLP autonomously adjusts weights to emphasize salient features and diminish less important ones. 
This capability is vital given the nuanced nature of mental health indicators and the potential for large datasets to contain a mix of relevant and irrelevant information. 

At the heart of our predictive model are two critical scores: the \texttt{sds score} and the \texttt{stai-s score}, representing depression and anxiety levels, respectively. Our primary objective was to develop a model that could accurately forecast these scores based on a comprehensive array of variables present in the dataset. These scores were derived through a sophisticated equation incorporating twenty distinct variables.
% 
% TODO -  can you please rewrite the equation using math symbols? It is hard to ascertain what these [stais1] etc. are? The paper will appear lot more professional with latex math equations. Please see here for a guide: https://www.overleaf.com/learn/latex/Mathematical_expressions
We followed the standard scoring procedures for each instrument to calculate the \texttt{stai-s score} and the \texttt{sds score}, as derived by \cite{shuster2021emotional}.

% OS: commented because we may not need these equations per the comment above.
% \begin{align}
% \text{\texttt{stai-s score}} = & (5 - \text{\texttt{sds score}}_1) + (5 - \text{STAI}_2) + \text{STAI}_3 + \text{STAI}_4 + \\
%                       & (5 - \text{STAI}_5) + (5 - \text{STAI}_6) + \text{STAI}_7 + (5 - \text{STAI}_8) + \\
%                       & \text{STAI}_9 + (5 - \text{STAI}_{10}) + (5 - \text{STAI}_{11}) + \text{STAI}_{12} + \\
%                       & \text{STAI}_{13} + \text{STAI}_{14} + (5 - \text{STAI}_{15}) + (5 - \text{STAI}_{16}) + \\
%                       & \text{STAI}_{17} + \text{STAI}_{18} + (5 - \text{STAI}_{19}) + (5 - \text{STAI}_{20})
% \end{align}
% \label{eq:stai_score}

% % Equation for \texttt{sds score}
% Similarly, the Self-Rating Depression Scale (SDS) score is calculated as:
% \begin{align}
% \text{\texttt{sds score}} = & \text{SDS}_1 + (5 - \text{SDS}_2) + \text{SDS}_3 + \text{SDS}_4 + \\
%                    & (5 - \text{SDS}_5) + (5 - \text{SDS}_6) + \text{SDS}_7 + \text{SDS}_8 + \\
%                    & \text{SDS}_9 + (5 - \text{SDS}_{10}) + (5 - \text{SDS}_{11}) + \text{SDS}_{12} + \\
%                    & \text{SDS}_{13} + \text{SDS}_{14} + (5 - \text{SDS}_{15}) + (5 - \text{SDS}_{16}) + \\
%                    & \text{SDS}_{17} + \text{SDS}_{18} + (5 - \text{SDS}_{19}) + (5 - \text{SDS}_{20})
% \end{align}
% \label{eq:sds_score}

\subsubsection{COVID-19 Depression and Anxiety Predictor (CoDAP) Architecture}
\label{sec:architecture}

\begin{wrapfigure}{r}{0.6\textwidth} % 'r' for right, 'l' for left, and '0.5\textwidth' is the width of the wrapfigure
  \centering
  % \vspace{-2ex}
  \includegraphics[width=\linewidth]{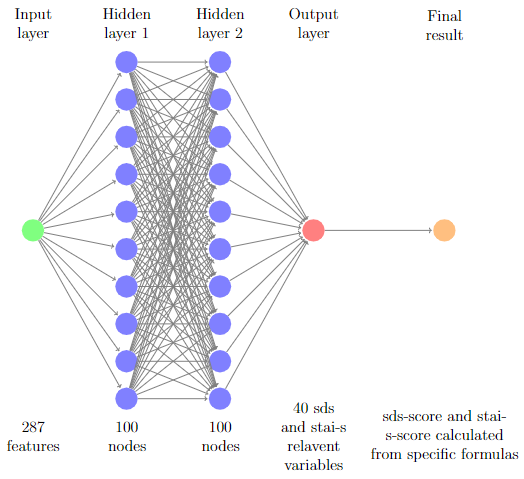}
  \vspace{-5ex}
  \caption{The CoDAP Architecture}
  % \vspace{-2ex}
  \label{fig:ann}
\end{wrapfigure}

Following our data preparation, we developed an MLP configured to leverage the entire spectrum of features in the dataset, thus enabling precise predictions of the \texttt{sds score} and \texttt{stai-s score}. The model architecture, as seen in Figure  \ref{fig:ann}, was designed to accommodate the complex relationships inherent in mental health data, focusing on capturing subtle patterns and indicators that less sophisticated models might miss. 
The input layer is marked by a green node, which encapsulates a comprehensive set of 287 unique attributes, excluding target attributes employed to calculate anxiety and depression levels, derived from the dataset by \citeauthor{shuster2021emotional}. 
These attributes form the foundational data that CoDAP will process. Some of these features include age, gender, race, religion, types of mental health disorders, and other qualitative features about the participants.
% TODO: please give some examples of the feature names (you can select some of the prominent feature names)

Careful consideration was given to the MLP network's architecture to balance predictive power and model generalizability.
It is a fundamental requirement for an MLP to have at least one hidden layer, which enables the model to perform non-linear transformations and learn complex relationships in the input data.
During the exploratory phase of our research, we conducted a series of experiments to determine the optimal number of hidden layers. Our empirical investigations revealed that a two-layered hidden structure provided the most robust performance. 
% We have two intermediate hidden layers as we progress deeper into the network's architecture. 

In the CODAP model architecture, each layer comprises 100 neurons, illustrated as blue nodes. These neurons are strongly connected, with each neuron in one layer forming connections to every neuron in the subsequent layer. 
The initial hidden layer acts as a feature transformer, mapping raw inputs into an intermediate space, and the second hidden layer builds upon this transformed feature space, enabling the MLP to integrate these initial patterns into higher-order representations. 
To enhance the network's ability to model non-linear relationships, a characteristic essential for capturing the complexities of mental health data, the Rectified Linear Unit (ReLU) activation function was applied within these hidden layers.

The output layer, indicated by red nodes, does not directly output the final \texttt{sds score} and \texttt{stai-s score}. Instead, it produces an intermediary set of 20 features to calculate these mental health scores. The output includes estimates for a series of scores: $STAI_1$ through $STAI_{20}$ and $SDS_1$ through $SDS_{20}$. Each set comprises 20 distinct scores. The \texttt{stai-s score}s gauge levels of anxiety, ranging from one (indicating minimal presence) to four (indicating a high degree), based on responses to various questions like ``I feel calm," ``I feel secure," and others along these lines.

Similarly, the \texttt{sds score}s measure aspects of depression on the same scale of one to four, with questions geared towards identifying symptoms of depression, such as changes in weight or eating habits, exemplified by statements like ``I notice I am losing weight" and ``I eat as much as I used to."

% TODO - please give some feature names from the 20 features in the output layer (again, you can select some of the prominent feature names)
The actual computation of the \texttt{sds score} and \texttt{stai-s score}, symbolized by yellow nodes, is performed using specific formulas derived from \cite{shuster2021emotional}. The output features include predictions of responses to each item within the questionnaires, which are then used to calculate the total scores for each scale.

The Adam optimization algorithm was employed to refine the network's predictive accuracy. This optimizer iteratively adjusts the network's weights to minimize the discrepancy, quantified by the Mean Squared Error (MSE), between CoDAP's predictions and the actual observed values from the dataset. 

\subsubsection{Cross Fold Validation}
\label{sec:cross-fold-validation}

% \begin{algorithm}
% \caption{K-fold Cross Validation with our ANN}
% \begin{algorithmic}[1]
% \Require
%   \Statex $data$: the dataset for cross-validation
%   \Statex $k$: the number of folds
%   \Statex $model$: the ANN model described in Figure \ref{fig:ann}
% \Ensure 
%     \Statex $AveragePerformance$: the average performance metric across all folds
%     \Statex
% \State $k\_folds \gets EquallyDivide(data)$
% \State $OverallPerformance \gets 0$
% \For{$i \gets 1$ to $k$}
%   \State $ValidationSet \gets k\_folds(i)$
%   \State $TrainingSet \gets k\_folds(all folds except i)$
%   \State $model.train(TrainingSet)$
%   \State $Performance \gets model.evaluate(ValidationSet)$
%   \State $OverallPerformance \gets OverallPerformance + Performance$
% \EndFor
% \State $AveragePerformance \gets OverallPerformance  /  k$
% \State \textbf{Output} $AveragePerformance$
% \end{algorithmic}
% \label{alg:cross-validation}
% \end{algorithm}

\begin{wrapfigure}{r}{0.75\textwidth} % 'r' for right and '0.5\textwidth' for the width of the wrapfigure
\vspace{-3em}
\begin{minipage}{\linewidth}
\begin{algorithm}[H]
\caption{K-fold Cross Validation with the CODAP model}
\begin{algorithmic}[1]
\Require
  \Statex $data$: the dataset for cross-validation
  \Statex $k$: the number of folds
  \Statex $model$: the MLP model described in Figure \ref{fig:ann}
\Ensure 
    \Statex $AveragePerformance$: the average performance metric across all folds
    \Statex
\State $k\_folds \gets EquallyDivide(data)$
\State $OverallPerformance \gets 0$
\For{$i \gets 1$ to $k$}
  \State $ValidationSet \gets k\_folds(i)$
  \State $TrainingSet \gets k\_folds(all folds except i)$
  \State $model.train(TrainingSet)$
  \State $Performance \gets model.evaluate(ValidationSet)$
  \State $OverallPerformance \gets OverallPerformance + Performance$
\EndFor
\State $AveragePerformance \gets OverallPerformance  /  k$
\State \textbf{Output} $AveragePerformance$
\end{algorithmic}
\label{alg:cross-validation}
\end{algorithm}
\end{minipage}
\vspace{-2ex}
\end{wrapfigure}

% To provide clarity on our methodology, we have included detailed pseudo-code illustrating the ANN's structure and function. This inclusion is intended to offer readers a clear understanding of the model's architecture and operational logic. 

A critical aspect of our methodology was the validation and optimization of our model. We utilized the K-fold cross-validation technique as a robust strategy to counteract overfitting and assure our model's reliability and generalizability across different subsets of the data demonstrated in \Cref{alg:cross-validation}. This method involved partitioning the dataset into several folds, systematically using each as a validation set while training the model on the remaining data. This allowed us to rigorously test and refine the model, ensuring its accuracy and robustness. 
This is crucial in mental health prediction, where overfitting to a particular subset of data could lead to misleading results.
Also, since each fold serves as a validation set at some point, cross-validation reduces the likelihood of the model's performance being biased by a particular random split of training and test data. This is especially important when dealing with complex and potentially heterogeneous datasets, such as those involving mental health symptoms during a pandemic.
Additionally, this method maximizes the data utilization, as the available data was limited. K-fold cross-validation allows for the efficient use of data, as each observation is used for training and validation at some point. 

\subsubsection{Hyperparameter Tuning}
\label{sec:hypermarameter}

For k-fold cross-validation, we opted for $k=5$ to strike a balance between obtaining a stable estimate of model performance and maintaining efficient computational processing. 
Additionally, we experimented with various hyperparameter tuning configurations of hidden layers, neurons per layer, and learning rates, each assessing the impact on model performance. This process was crucial for identifying the optimal parameters to produce the most accurate and reliable predictions. 
As a result, the hyperparameters specified for our model, CoDAP, are as follows: learning rate of 0.01, 100 epochs of training, ReLU activation function, Adam optimization algorithm, and the MSE as the loss function.
% TODO - are you able to produce some graphs of MSE with these various values changed? i.e., MSE with different learning rate, different epochs, etc? (I would not actually call ``two hidden layers, 287 nodes in the input layer, 100 nodes in each hidden layer, two nodes in the output layer" as hyperparamters, as those are not necessarily tunable! If you experimented with different number of layers in the model architecture, then we can add something about that.
% \begin{table}[h]
%     \centering
%     \caption{MSE with Different Learning Rates}
%     \setlength{\tabcolsep}{1.2mm}{
%     \begin{tabular}{lcccrrr}
%     \toprule
%         \textbf{Learning Rate} & \textbf{0.1} & \textbf{0.05} & \textbf{0.01} & \textbf{0.005} & \textbf{0.001} \\ \midrule
%         MSE & 2.8502 & 1.5652 & 0.5239 & 0.5630 & 0.5939 \\
%     \bottomrule
%     \end{tabular}}
%     \label{tab:mse-lr}
% \end{table}

% \begin{table}[h]
%     \centering
%     \caption{MSE with Different Hidden Layer Size (lr = 0.01)}
%     \setlength{\tabcolsep}{2mm}{
%     \begin{tabular}{lcccrrr}
%     \toprule
%         \textbf{nodes} & \textbf{10} & \textbf{50} & \textbf{100} & \textbf{150} & \textbf{200} \\ \midrule
%         MSE & 0.8576 & 0.5474 & 0.5420 & 0.5713 & 0.6234 \\
%     \bottomrule
%     \end{tabular}}
%     \label{tab:mse-layer}
% \end{table}

\begin{table}[h]
    \centering
    \begin{minipage}[t]{0.5\linewidth}
        \centering
        \setlength{\tabcolsep}{0.5mm} % adjust the space between columns as needed
        \begin{tabular}{lccccc}
        \toprule
            \textbf{LR} & \textbf{0.1} & \textbf{0.05} & \textbf{0.01} & \textbf{0.005} & \textbf{0.001} \\ \midrule
            MSE & 2.8502 & 1.5652 & 0.5239 & 0.5630 & 0.5939 \\
        \bottomrule
        \end{tabular}
        \caption{MSE with Different Learning Rates}
        \label{tab:mse-lr}
    \end{minipage}% The percent sign removes the space that would otherwise appear between the minipages
    \begin{minipage}[t]{0.5\linewidth}
        \centering
        \setlength{\tabcolsep}{0.5mm} % adjust the space between columns as needed
        \begin{tabular}{lccccc}
        \toprule
            \textbf{Nodes} & \textbf{10} & \textbf{50} & \textbf{100} & \textbf{150} & \textbf{200} \\ \midrule
            MSE & 0.8576 & 0.5474 & 0.5420 & 0.5713 & 0.6234 \\
        \bottomrule
        \end{tabular}
        \caption{MSE with Different Layers (lr = 0.01)}
        \label{tab:mse-layer}
    \end{minipage}
    \vspace{-3ex}
\end{table}

Upon finalizing the hyperparameters, we trained CoDAP using the entire training set, which incorporated all the folds from our cross-validation process. 
The model thus trained was then employed to predict and compare results with the designated testing set. This approach, encompassing multiple rounds of training, hyperparameter adjustments, and validation, contributed significantly to the robustness and reliability of CoDAP for predictive analysis. 

\subsection{Explaining the Model Results}
\label{sec:lime}

% The complexity of our CoDAP model poses a huge challenge in comprehending the rationale behind specific predictions generated by the model. 
To explain the CoDAP model results, we employed Local Interpretable Model-agnostic Explanations (LIME)~\cite{ribeiro2016should} for approximating the model with a simpler, interpretable model around the vicinity of the prediction that needs to be explained. LIME is particularly useful because it is model-agnostic.
% as a tool to establish locally explainable models that approximate the intricate functioning of the complex model within the proximity of a given input instance. 

\begin{figure}[h!]
  \centering
  \includegraphics[width=\textwidth]{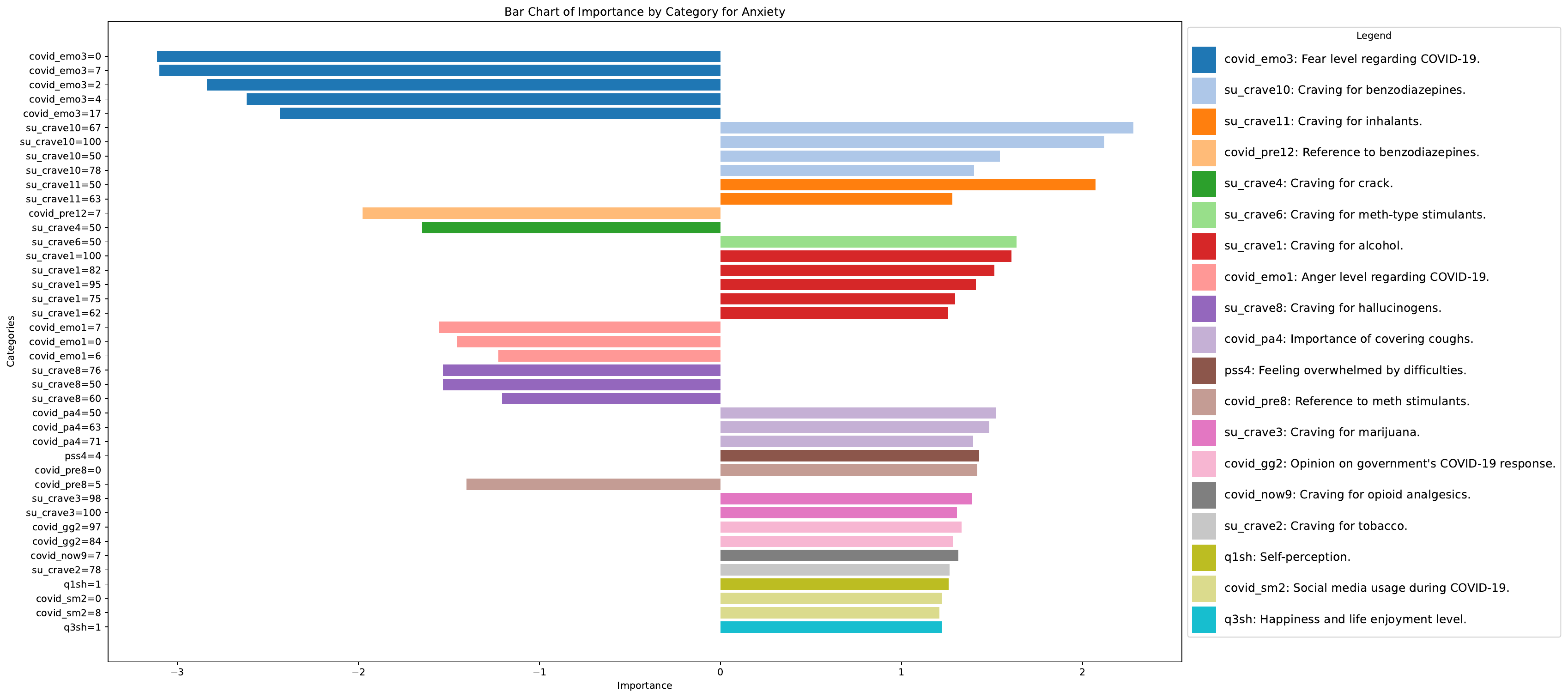}
  \vspace{-3ex}
  \caption{LIME \texttt{stai-s score} (Anxiety) Result}
  \vspace{-2ex}
  \label{fig:stais_lime_result}
\end{figure}

\begin{figure}[h!]
  \centering
  \includegraphics[width=\textwidth]{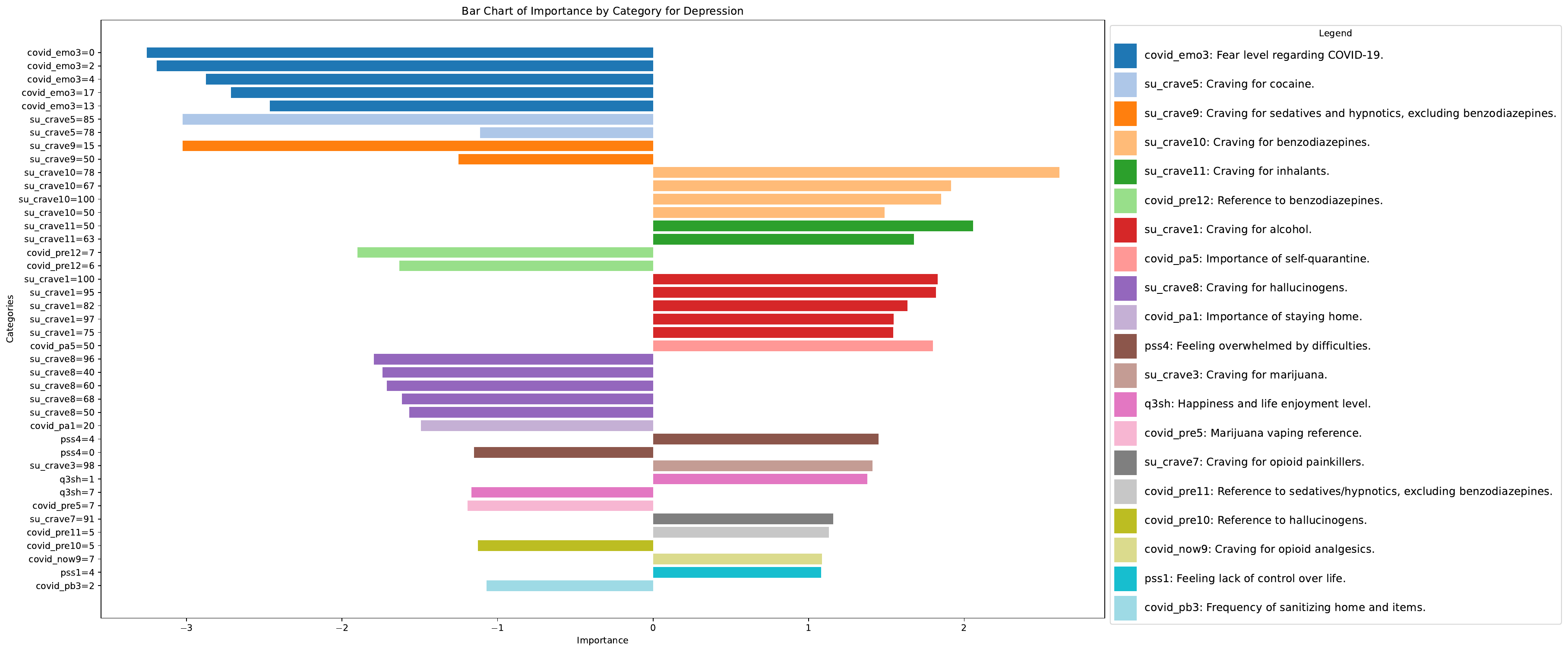}
  \vspace{-4ex}
  \caption{LIME \texttt{sds score} (Depression) Result}
  \label{fig:sds_lime_result}
  % \vspace{-2ex}
\end{figure}

Because LIME is designed to explain model predictions for individual instances rather than offering a comprehensive interpretation of the entire model, we created an objective overview of the most important features.
We first normalized the values for each of our features. Then, we leveraged LIME's local interpretation, dividing individual features into categories specified by LIME and averaging importance scores. Afterward, we carried out a series of ten iterative steps. We generated LIME reports in each iteration that focused separately on two specific scores: \texttt{stai-s score} and \texttt{sds score}, based on our test dataset. After the ten iterations, we grouped the same features and averaged their importance scores. This process helped us understand how each category of features influenced the \texttt{stai-s score} and \texttt{sds score} across all iterations. We then created a graphical representation illustrating the most important categories corresponding to both \texttt{stai-s score} and \texttt{sds score}. In this step, a uniform color scheme was adopted for categories belonging to the same feature to improve the readability of the graph. We selected the top 40 most important features from there and grouped items within the same category. This approach helped discern the important features that influence a particular prediction to explain the predictive capabilities of our model.
% , thereby fostering a sense of confidence and trust in the predictive capabilities of our model.

In particular, by averaging the influence scores generated by LIME, we identified key features that significantly impact the model's predictions. Notably, certain features exerted either a positive or negative influence on the model's anxiety and depression scores. We started by recognizing the category using the legend and finding the corresponding y-axis score to interpret the graph. The x-axis displays the overall importance score from LIME, where negative values indicate a negative correlation with the anxiety or depression scores, and positive values suggest a positive correlation with the anxiety or depression scores.

Among the most influential features were \texttt{su\_crave10} and \texttt{su\_crave3}, which gauge the craving intensity for substances like Xanax and marijuana. Both of these features showed a positive correlation with anxiety, aligning with the understanding that individuals with heightened cravings for addictive substances tend to exhibit increased stress levels. This can be correlated with two alternative explanations due to the COVID-19 pandemic: 1) supply for these drugs was disrupted, leading to increased mood symptoms; 2) people who were taking these already had more mental health issues to start with; therefore, it is not surprising they craved more for drugs during the pandemic and had more depression/anxiety.

Another notable feature, \texttt{Covid\_emo3}, measures the fear associated with COVID-19. Interestingly, this feature displayed a negative correlation with anxiety and depression at lower levels, suggesting that individuals with diminished COVID-19 fears might experience less overall stress and, subsequently, fewer mood symptoms. This observation is consistent with the premise that fear of COVID-19 is a primary stressor during this period. Similarly, additional features, such as \texttt{Covid\_emo4}, demonstrated parallel trends, reinforcing our model's ability to capture the nuanced interplay between various factors and their impact on anxiety and depression.

\section{Results}
\label{sec:results}

% In our evaluation, we utilized graphs to visually represent the findings from our data analysis and the results of our Artificial Neural Network (ANN). These graphical representations included trends in anxiety and depression scores over time, correlations between different qualitative attributes, and the performance of our ANN model in predicting these scores. 

Our exploratory data analysis revealed noteworthy patterns and correlations. We observed trends in mental health scores across different demographic groups and during various pandemic phases. 
For instance, our findings (see \Cref{fig:heatmap_correlation}) indicate that across all participants and pandemic phases, younger people typically exhibited higher levels of depression and anxiety. Moreover, there was a negative correlation between income and the levels of depression and anxiety, with individuals of lower income often experiencing more pronounced mental health challenges during the pandemic period.
These insights were instrumental in shaping the design and focus of CoDAP, providing a solid foundation for predictive analysis. 

% The performance of our model, CoDAP, was evaluated using MSE. 
% TODO: did you actually calculate  precision and recall? They are not presented in the results below. So, please either add to the results or remove the mention of those metrics above.
 Since predicting the level of anxiety and depression can be framed as a regression problem, we utilized the MSE, which was particularly suitable for this task.
The CODAP model demonstrated high accuracy in predicting the \texttt{sds} and \texttt{stai-s scores}, indicating its effectiveness in capturing the complex relationships within the data. The use of MSE as a loss function was particularly beneficial, given its sensitivity to large errors and computational efficiency. The MSE also provided a standard metric to compare the performance of different models and configurations.
Table \ref{tab:univ-compa} presents the MSE in the five rounds of the cross-validation process.

% \begin{table}[h]
%     \centering
%     \caption{MSE for Cross-Validation}
%     \setlength{\tabcolsep}{0.6mm}{
%     \begin{tabular}{lcccrrr}
%     \toprule
%         \textbf{Round} & \textbf{1} & \textbf{2} & \textbf{3} & \textbf{4} & \textbf{5} & \textbf{Average} \\ \midrule
%         MSE & 0.5064 & 0.4924 & 0.5720 & 0.5401 & 0.5393 & \textbf{0.5300} \\
%     \bottomrule
%     \end{tabular}}
%     \label{tab:univ-compa}
% \end{table}

\begin{table}[h]
    \centering
    \setlength{\tabcolsep}{0.6mm}{
    \begin{tabular}{|l|c|c|c|c|c|r|} % Added vertical bars to all columns, and double bars before the "Average"
    \hline % Use \hline for horizontal lines
    \textbf{Round} & \textbf{1} & \textbf{2} & \textbf{3} & \textbf{4} & \textbf{5} & \textbf{Average} \\
    \hline
    MSE & 0.5064 & 0.4924 & 0.5720 & 0.5401 & 0.5393 & \textbf{0.5300} \\
    \hline
    \end{tabular}}
    \caption{MSE for Cross-Validation}
    \label{tab:univ-compa}
    \vspace{-2ex}

\end{table}

Throughout the training and k-fold cross-validation process, the CODAP model predicts 20 features each for calculating \texttt{sds score} and \texttt{stai-s score}. Despite the limitation of a relatively small data set, the model's predictive performance is promising, with an average MSE of approximately 0.53 across five rounds of cross-validation for the feature values utilized in computing the \texttt{sds score} and \texttt{stai-s score}.
% with 1 denoting the minimum level and 4 representing the maximum level.
In the dataset, the feature values were assigned positive or negative weights based on their contributions to increasing or decreasing levels of anxiety and depression. The outcome features \texttt{sds score} and \texttt{stai-s score} were normalized to range from 1 to 4, with 1 denoting the minimum level and 4 representing the maximum level. 
Also, the cross-validation outcomes affirm the model's reliability and underline its robustness, with MSE values demonstrating consistency within the range of 0.49 to 0.57 across different training rounds. 
This consistency is crucial because it suggests that the model is stable and reliable across different subsets of data.
Although a smaller MSE would be preferred, given the primary task of the MLP is to understand which features are important predictors of mental health state and the range and the variance of the outcome features, i.e., the target variables, we deem these average MSE of 0.53 to be a good metric for the task at hand.

\section{Conclusion and Future Work}

Mental health has become an increasingly important area in healthcare, and this trend is anticipated to persist, if not escalate, in the future. As such, being able to predict mental health symptoms is critical for us to monitor and intervene, given the rising level of global health crises, including and beyond COVID-19. 
% 
% The methodologies and procedures employed in our research were carefully designed and validated to ensure their rigor and reliability. % OS: commented because it is superfluous
Our exploratory data analysis, data reformatting, and robust model validation techniques contribute to the validity of our research findings. 
% The choice of MSE as our loss function aligned with the model's objectives and the practical constraints of our intended psychiatry applications. 
The model's performance suggests a promising foundation, even though it is based on a limited data set. 
% However, the primary intention of our research was to utilize this dataset to understand resilience and emotional recovery metrics in the context of a global traumatic event. 
By focusing on the quantitative attributes of individuals (e.g., demographics, COVID-specific behaviors, etc.) and the time elapsed since the event, our research provides valuable insights into the dynamics of mental health during such crises. 
Our model can be tailored and refined with additional data and feedback. This ongoing improvement is particularly important in a rapidly changing scenario like a pandemic, where initial models may need refinement as more data becomes available and the nature of the mental health challenges evolves.
However, since the CoDAP's training dataset was primarily collected during the COVID-19 pandemic~\cite{shuster2021emotional}, the results may not generalize well to pre- and post-pandemic conditions. 
% 
% Our long-term goal is to extend the applicability of our findings to other global traumatic events beyond COVID-19. 
% 
Nevertheless, we believe the CODAP model can be adapted and applied to different contexts, providing valuable tools for understanding and predicting mental health responses to various environmental and chronic stressors. 
% We are expecting enhanced results as the model undergoes training with larger data sets in the future.
% 
% Our vision for CoDAP is to translate the insights generated by the model into an application, enabling users to complete a detailed survey and providing valuable data about their mental health over time. %More importantly, it would offer insights into their current and future mental health status, including resilience and emotional recovery metrics. 

Our vision for CoDAP is to harness the model's predictive power by integrating it into a user-friendly application that will capture baseline mental health measures, including but not limited to age, income, and craving — features our model identified as having significant predictive importance. The application will then be able to forecast future changes in mental health status based on these baseline measures and ongoing user interaction. Consequently, while direct measures of depression and anxiety are integral components of the initial assessment, the predictive model will be essential for providing longitudinal insights and anticipating future mental health trends for users. This approach aims to offer users not only a snapshot of their current mental health status but also a proactive tool for monitoring and managing their mental health over time.
% OS: TODO commented the future work part to reclaim some space 
To achieve this, we plan to utilize the Auto-Regressive Integrated Moving Average (ARIMA) model~\cite{nelson1998time} for time series prediction. 
% This approach will allow us to effectively monitor and analyze key indicators such as the \texttt{sds score} (depression) and the \texttt{stai-s score} (anxiety) among individuals over time. 
Such longitudinal analysis is critical for understanding the dynamics of mental health and identifying patterns or changes that may require intervention. 
% Also, often, the stigma surrounding mental health leads users to prioritize the privacy and security of their data. Therefore, we plan to develop a decentralized model utilizing federated learning in response to this concern. This approach ensures that data stays confined to the user's device. Implementing a federated learning system will allow us to create a system that yields valuable insights into mental health and diligently safeguards and respects the sensitive nature of the information it processes. 

In conclusion, the ultimate goals of this work are to enhance understanding of mental health dynamics, using the predictive power of MLPs that analyze complex psychological data and extend to other techniques such as time-series analysis and transfer learning to gauge individuals' mental health conditions in wide-ranging scenarios.
% during both crises and non-crises situations.
% \noindent\textbf{Resources:} The anonymized version of our GitHub repository with our code and  documentation is available at https://anonymous.4open.science/r/MentalHealth-8BA6.

\bigskip

\noindent\textbf{Resources:} The  GitHub repository with our code and  documentation is available at \url{https://github.com/block-iot/MentalHealth}.
% In summary, our project is poised at the intersection of mental health research and technological innovation. By harnessing advanced predictive models and prioritizing user privacy, we aim to create a tool that is not only effective in monitoring mental health metrics but also accessible and secure for its users. % OS: commented because it is a bit redundant

\bibliographystyle{ACM-Reference-Format}
\bibliography{reference}

\end{document}